\documentclass[conference]{IEEEtran}
\IEEEoverridecommandlockouts

\usepackage{cite}
\usepackage{amsmath,amssymb,amsfonts}
\usepackage{algorithmic}
\usepackage{graphicx}
\usepackage{textcomp}
\usepackage{xcolor}
\def\BibTeX{{\rm B\kern-.05em{\sc i\kern-.025em b}\kern-.08em
    T\kern-.1667em\lower.7ex\hbox{E}\kern-.125emX}}
\usepackage{bbm}
\usepackage{makecell}
\usepackage{booktabs}
\usepackage{gensymb}
\usepackage[colorlinks=true, citecolor=blue, linkcolor=blue, urlcolor=blue]{hyperref}
    
\begin{document}

\title{CRC-Router: Risk-Constrained Routing for Medical Agentic AI Systems}

\author{
\IEEEauthorblockN{
Xueyang Li,
Mingze Jiang,
Gelei Xu,
Jun Xia,
Ching-Hao Chiu,
Mengzhao Jia,
Danny Z. Chen,
and Yiyu Shi
}
\IEEEauthorblockA{
Computer Science and Engineering, University of Notre Dame, USA \\
\{xli34, mjiang23, gxu4, jxia4, cchiu3, mjia2, dchen, yshi4\}@nd.edu
}
}

\maketitle

\begin{abstract}
Agentic AI systems are increasingly being explored in medical imaging to improve throughput and reduce clinician workload; however, safe deployment remains challenging because autonomous errors may propagate into downstream clinical decisions. A central requirement is therefore not only strong predictive performance, but also a reliable routing mechanism that determines when the system should proceed autonomously and when a case should be escalated for further review. To address this gap, we propose CRC-Router, a risk-constrained, uncertainty-aware routing module that is applicable to both conventional medical prediction models and agentic medical AI systems. CRC-Router combines multiple complementary uncertainty signals with the predictive score to construct a per-finding routing feature vector, maps this vector to an estimated wrong-accept risk using a lightweight per-finding risk model, and then applies Conformal Risk Control (CRC) to calibrate acceptance thresholds under a user-specified risk target. Instantiated on chest X-ray multi-finding triage using the NIH ChestX-ray14 dataset, CRC-Router achieves the strongest empirical risk--coverage trade-off among the evaluated baselines, both as a standalone routing layer and as a plug-in module integrated with the state-of-the-art MedRAX agent. These results demonstrate both the effectiveness of CRC-Router in selective medical automation and its modular, model-agnostic compatibility with existing predictive and agentic medical pipelines. Code is publicly available at \href{https://github.com/XLIAaron/CRC-Router}{https://github.com/XLIAaron/CRC-Router}.
\end{abstract}


\section{Introduction}
Agentic AI systems have attracted substantial attention for their planning, self-adjustment, and autonomous decision-making capabilities. In medical imaging, they show promise for improving clinical throughput and reducing workload~\cite{xu2025comprehensive}. However, increased autonomy raises the impact of 
errors: failures may propagate beyond isolated predictions to downstream clinical decisions. This is especially critical in healthcare, where model outputs can affect diagnosis and treatment, and errors can cause delayed care, financial burden, or even patient harm. Thus, clinical agentic AI systems must not only achieve strong discrimination, but also operate as responsible AI systems that explicitly account for risks and support safe and transparent deployment.

In practice, this safety requirement is tightly connected to uncertainty estimation, as uncertainty signals often provide the most accessible evidence for when autonomous predictions are likely to fail. Although uncertainty-aware mechanisms have been explored in agentic AI, primarily in general-domain language and tool-use settings, these methods typically use uncertainty as a heuristic coordination signal (e.g., for tool orchestration~\cite{han2024towards}, trajectory propagation~\cite{zhao2025uncertainty}, adaptive intervention~\cite{zhi2025seeing,zhang2026agentic}). Such uses are valuable for improving agent behavior, but they do not directly yield a formally calibrated routing policy with explicit risk control. Consequently, they do not address the central deployment requirement in our setting: enforcing a user-specified risk constraint on accept-versus-escalate decisions through an explicitly risk-calibrated routing rule.

A natural candidate for enforcing risk constraints is Conformal Risk Control (CRC)~\cite{angelopoulos2022conformal}, which provides statistically grounded calibration for black-box systems. While it has shown value in medical imaging~\cite{silva2025trustworthy,teneggi2025conformal}, prior works focus strictly on task-level or prediction-specific risk. However, medical agentic AI systems require a conceptual shift toward treating routing as a distinct, decision-theoretic layer above prediction. This routing determines whether to execute a prediction autonomously or escalate it, explicitly controlling which errors propagate into downstream clinical workflows. Such rigorous control fundamentally differs from selective prediction~\cite{geifman2017selective} and its reliance on heuristics without formal risk guarantees. To the best of our knowledge, CRC remains unstudied as a routing-layer calibration mechanism for risk-constrained decisions in medical agentic AI systems. Indeed, applying it requires overcoming nontrivial design challenges: determining how to represent heterogeneous uncertainty, convert it into risk estimates, and integrate this routing layer into existing agentic pipelines.

To address this gap, we propose \textbf{CRC-Router, a novel formalization of risk-constrained routing for both conventional predictive models and medical agentic AI systems with statistical wrong-accept guarantees.} Unlike selective classification, which merely rejects uncertain predictions, CRC-Router explicitly calibrates wrong-accept risk under operational coverage constraints. It maps multi-signal uncertainty features and predictive scores to a per-finding risk estimate, then uses CRC to calibrate actionable routing thresholds under a user-specified safety target. Crucially, this design decouples predictive modeling from autonomy control, yielding a model-agnostic, plug-in compatible module. We instantiate this framework on chest X-ray multi-finding triage, evaluating it as a standalone routing layer and as a plug-in for the MedRAX agent~\cite{fallahpour2025medrax}. Across both settings, CRC-Router achieves the best risk--coverage trade-off among evaluated methods, demonstrating that it provides effective risk control and is compatible with existing medical agentic AI systems.



\section{Background}
\label{sec:background}

\subsection{Medical Agentic AI and Selective Automation}

Medical AI systems are increasingly adopting agentic designs in which large language or multimodal models reason over intermediate states, invoke external tools, and coordinate multi-step decision processes rather than producing a single direct prediction. Representative examples include MDAgents~\cite{Kim2024MDAgents}, which studies adaptive collaboration among multiple LLMs for medical decision-making, and MMedAgent~\cite{Li2024MMedAgent}, which learns to select specialized medical tools across tasks and modalities within a unified multimodal framework. Similar designs have also emerged in medical imaging, where systems such as MedRAX~\cite{fallahpour2025medrax} and RadFabric~\cite{Chen2025RadFabric} compose perception modules, diagnostic tools, and multimodal reasoning components for CXR interpretation. Together, these works show a broader shift from isolated medical predictors toward orchestrated, tool-using AI systems.

The closest classical literature is selective prediction, where a model may reject unreliable examples to improve the risk--coverage trade-off~\cite{geifman2017selective}. However, selective prediction does not fully align with medical agentic AI: rejection is usually treated as a terminal outcome, whereas medical deployment requires an intermediate routing decision to a stronger model, downstream agent, or human expert. Moreover, most selective-prediction methods rely on confidence scores or learned rejection functions rather than an explicitly calibrated \emph{wrong-accept} risk. Selective prediction therefore provides an important foundation, but not a complete solution, for risk-constrained routing in selective medical automation.

\subsection{Uncertainty Estimation for Risk-Aware Routing}

Predictive confidence, often measured by the maximum predicted class probability, provides a natural starting point for identifying unreliable predictions. However, confidence alone is an incomplete routing signal: modern neural networks are often poorly calibrated, so high softmax confidence does not necessarily imply a high probability of correctness~\cite{Guo2017Calibration}. More severely, deep networks can assign near-certain confidence to unrecognizable or far-out-of-distribution inputs, implying that confidence-only routing may accept cases that should instead be escalated. Predictive entropy, rooted in information-theoretic uncertainty quantification, provides a posterior-level summary of uncertainty and has become a standard signal for failure detection and selective prediction~\cite{GalGhahramani2016}. Nevertheless, entropy remains derived from the predictive distribution alone and cannot distinguish whether uncertainty arises from model disagreement, intrinsic input ambiguity, or distributional shift.


In Bayesian deep learning, epistemic uncertainty reflects uncertainty in the model itself, such as limited training support, and can be approximated using Monte Carlo dropout~\cite{GalGhahramani2016}. Aleatoric uncertainty instead reflects ambiguity intrinsic to the observation, such as noise, occlusion, or equivocal visual evidence~\cite{KendallGal2017}. A complementary perspective is distributional atypicality: posterior-based uncertainty may remain overconfident under dataset shift~\cite{Ovadia2019}, whereas Mahalanobis-style scoring measures deviation from in-distribution feature statistics~\cite{Lee2018Mahalanobis}. These observations motivate integrating predictive entropy, epistemic and aleatoric uncertainty, and distributional scoring as complementary routing evidence rather than relying on a single confidence score.


\subsection{Conformal Calibration and Risk Control}

Conformal prediction offers a model-agnostic, distribution-free framework for uncertainty quantification under exchangeability~\cite{angelopoulos2023conformal}. In classification and structured prediction, its natural output is a set or interval that enjoys formal finite-sample coverage guarantees. This is highly valuable for reliability, but it does not by itself resolve the selective automation problem studied here. In our setting, the deployment question is not only how to represent uncertainty, but how to translate uncertainty into an actionable decision about whether autonomous execution should proceed. A set-valued output therefore still requires an additional decision rule before it can govern accept-versus-escalate behavior in an agentic workflow.

Conformal Risk Control extends conformal ideas beyond miscoverage to bounded monotone losses, allowing decisions to be calibrated against a user-specified expected-risk target~\cite{angelopoulos2022conformal}. Medical-imaging applications of conformal calibration include semantically adaptive uncertainty quantification in CT~\cite{teneggi2025conformal} and few-shot transfer of medical vision-language models~\cite{silva2025trustworthy}. Here, we apply CRC to the \emph{wrong-accept risk}, defined as the probability that a finding prediction is both accepted and incorrect, to calibrate per-finding accept-versus-escalate decisions.

The next section formalizes this routing problem and instantiates a CRC-based procedure for calibrating accept-versus-escalate decisions under a user-specified target risk.

\begin{figure*}[t]
    \centering
    \includegraphics[width=\textwidth]{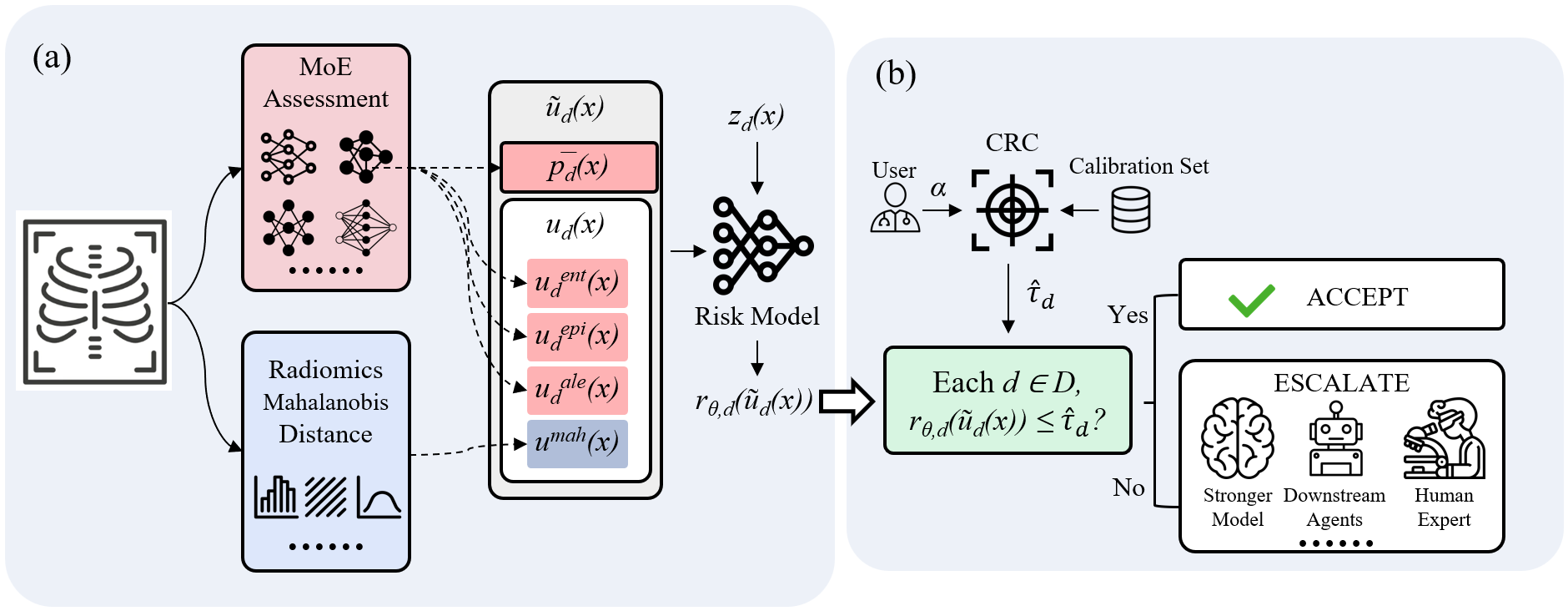}
    \caption{Overview of the proposed CRC-Router. (a) Multi-signal uncertainty and risk modeling. (b) Conformal Risk Control (CRC)-based threshold calibration and deployment routing.}
    \label{fig:framework}
\end{figure*}

\section{Method}
\label{sec:method}

\subsection{Problem Formulation}
\label{subsec:problem_formulation}

Let \(x \in \mathcal{X}\) denote an input image, and let \(y=(y_1,\dots,y_D)\in\{0,1\}^D\) denote the multi-label ground-truth vector over \(D\) findings. For each finding \(d\in\{1,\dots,D\}\), a base model (or ensemble) produces a predictive score \(\bar p_d(x)\in[0,1]\), and a hard prediction \(\hat y_d(x)=\mathbbm{1}[\bar p_d(x)\ge t_d]\), where \(t_d\) is a disease-specific decision threshold (e.g., fixed at \(0.5\) or tuned on a non-test split). The router then outputs an action \(a_d(x)\in\{\texttt{accept},\texttt{escalate}\}\). Here, \texttt{accept} means the system proceeds with the autonomous prediction for that finding, whereas \texttt{escalate} routes the case for additional review (e.g., a stronger model, a downstream agent, or a human expert). Because only accepted predictions are acted upon autonomously, the safety-relevant error for routing is the \textbf{accepted error}. Accordingly, we define the per-finding wrong-accept risk as \(R_d=\mathbb{P}\!\big(a_d(x)=\texttt{accept}\ \wedge\ \hat y_d(x)\neq y_d\big)\), which measures how often the system makes an autonomous incorrect decision on finding \(d\). We define coverage as \(\phi_d=\mathbb{P}\!\big(a_d(x)=\texttt{accept}\big)\), i.e., the fraction of cases handled autonomously. Our objective is to maximize coverage \(\phi_d\) subject to a per-finding risk constraint \(R_d\le \alpha\), where \(\alpha\) is a user-specified target risk level. In this work, we instantiate this risk-constrained routing formulation for chest X-ray (CXR) multi-finding triage as a case study. Under this setting, \(x\) denotes a chest X-ray image, and each index \(d\in\{1,\dots,D\}\) corresponds to a candidate finding (disease label) for that image. 

Fig.~\ref{fig:framework} summarizes the proposed CRC-Router. The framework comprises two stages: (a) multi-signal uncertainty feature construction with per-finding risk modeling, and (b) CRC-based threshold calibration for deployment-time accept-versus-escalate routing under a user-specified risk target. We describe each stage in the following subsections.

\subsection{Multi-Signal Uncertainty Vector}
\label{subsec:uncertainty_vector}

As illustrated in Fig.~\ref{fig:framework}(a), the first component of CRC-Router constructs a per-finding multi-signal uncertainty representation from committee predictions and an image-level distributional score, which is then used for downstream risk modeling. Formally, let \(\{p_{m,d}(x)\}_{m=1}^M\) denote the probability predictions for finding \(d\) produced by \(M\) committee members, where \(m\) indexes the committee member. From these committee outputs, we define the ensemble mean probability as \(\bar p_d(x)=\frac{1}{M}\sum_{m=1}^M p_{m,d}(x)\). For binary predictions, we use the entropy function \(H(p)=-p\log p-(1-p)\log(1-p)\). We then define a per-finding uncertainty vector \(u_d(x)\in\mathbb{R}^4\), whose components are described below.

\subsubsection{Predictive Entropy}
Predictive entropy captures overall output uncertainty from the aggregated committee prediction, but does not distinguish whether the uncertainty arises from model disagreement or intrinsic ambiguity in the input:
\begin{equation}
u^{\mathrm{ent}}_d(x)=H\!\big(\bar p_d(x)\big).
\label{eq:u_ent}
\end{equation}
This term serves as a compact posterior-level uncertainty summary and provides a strong baseline signal for selective prediction, but is insufficient on its own for risk-sensitive routing under distribution shift.

\subsubsection{Epistemic Uncertainty}
In contrast to predictive entropy, which summarizes uncertainty after aggregation, epistemic uncertainty is intended to capture disagreement across models. We estimate it using the variance of committee predictions:
\begin{equation}
u^{\mathrm{epi}}_d(x)=\mathrm{Var}_{m=1}^{M}\!\big(p_{m,d}(x)\big),
\label{eq:u_epi}
\end{equation}
where \(\mathrm{Var}_{m=1}^{M}(\cdot)\) denotes the empirical variance across committee predictions. A large value indicates that committee members disagree on the same input, suggesting elevated model uncertainty, such as limited support in training data, model instability, or sensitivity to representation differences.

\subsubsection{Aleatoric Uncertainty}
Whereas epistemic uncertainty measures between-model disagreement, aleatoric uncertainty characterizes uncertainty that appears within individual model predictions and is therefore more closely associated with data ambiguity, including noisy or visually ambiguous findings. We estimate this using the average entropy across committee members:
\begin{equation}
u^{\mathrm{ale}}_d(x)=\frac{1}{M}\sum_{m=1}^{M} H\!\big(p_{m,d}(x)\big).
\label{eq:u_ale}
\end{equation}
This complements \(u^{\mathrm{epi}}_d(x)\): two cases may have similar predictive entropy, but differ in whether uncertainty is driven by model disagreement (epistemic) or consistently uncertain per-model predictions (aleatoric).

\subsubsection{Mahalanobis-Style Distributional Score}
The preceding three terms are derived from model outputs. To additionally capture distributional atypicality that may not be reflected in posterior probabilities, we include a Mahalanobis-style feature-space discrepancy score, denoted \(u^{\mathrm{mah}}(x)\). Mahalanobis distance can be defined in various feature spaces; in this work, we compute it using handcrafted radiomics-style image descriptors (first-order intensity, texture, and histogram-shape features) rather than deep embeddings. This choice provides a model-agnostic distributional signal that is independent of the committee outputs and incurs minimal computational overhead, thereby preserving low-latency routing and avoiding unnecessary delay in autonomous agent execution. Specifically, we fit a diagonal Mahalanobis model on the training distribution in this handcrafted feature space, producing a single image-level score \(u^{\mathrm{mah}}(x)\) that is shared across all findings. This term provides an image-level signal for outlier-like or shifted inputs and complements posterior-based uncertainty measures.

Combining the above terms, the per-finding uncertainty vector is
\begin{equation}
u_d(x)=\Big[u^{\mathrm{ent}}_d(x),\;u^{\mathrm{epi}}_d(x),\;u^{\mathrm{ale}}_d(x),\;u^{\mathrm{mah}}(x)\Big]\in\mathbb{R}^4.
\label{eq:uncertainty_vector}
\end{equation}
Since the probability of error depends not only on uncertainty signals but also on the predictive score, we augment the multi-signal uncertainty vector \(u_d(x)\) with the predictive score \(\bar p_d(x)\). This yields a per-finding routing feature vector \(\tilde u_d(x)=[u_d(x),\bar p_d(x)]\in\mathbb{R}^5\), as shown in Fig.~\ref{fig:framework}(a). We then learn a per-finding risk model \(r_{\theta,d}\), defined by
\begin{equation}
r_{\theta,d}:\mathbb{R}^5\rightarrow[0,1], 
\qquad
r_{\theta,d}\!\big(\tilde u_d(x)\big)\approx \mathbb{P}\!\big(\hat y_d(x)\neq y_d \mid \tilde u_d(x)\big),
\label{eq:risk_model_routing}
\end{equation}
where \(\theta\) denotes the learnable parameters of the risk model, with a separate parameter set fitted for each finding \(d\); \(y_d\in\{0,1\}\) is the ground-truth label for finding \(d\), and \(\hat y_d(x)\) is the predicted label for input \(x\). The model is supervised using the binary error indicator \(z_d(x)=\mathbf{1}[\hat y_d(x)\neq y_d]\) on an out-of-sample calibration split. In practice, we use a lightweight MLP, since the uncertainty-to-risk mapping may be nonlinear and involve feature interactions, while the input dimension is small and low-latency execution is desirable for routing. 

\subsection{Conformal Risk Control for Threshold Calibration}
\label{subsec:crc}
As illustrated in Fig.~\ref{fig:framework}(b), the risk score produced by the uncertainty-to-risk model is subsequently converted into a deployable routing rule through a calibration step. Specifically, the risk model in Eq.~\eqref{eq:risk_model_routing} maps the routing feature vector \(\tilde u_d(x)\) to a per-finding risk score \(r_{\theta,d}(\tilde u_d(x))\), which estimates the conditional probability that the corresponding prediction \(\hat y_d(x)\) is incorrect. To instantiate the risk-constrained routing objective in Sec.~\ref{subsec:problem_formulation}, this score must be converted into a calibrated routing rule. We therefore adopt Conformal Risk Control (CRC)~\cite{angelopoulos2022conformal}, which calibrates a per-finding acceptance threshold on the predicted risk score so as to control the expected wrong-accept risk at a user-specified level \(\alpha\).

For a finding \(d\) and a candidate threshold \(\tau\), CRC induces the acceptance rule \(a_d^{(\tau)}(x)=\texttt{accept}\) iff \(r_{\theta,d}(\tilde u_d(x))\le \tau\), and \(a_d^{(\tau)}(x)=\texttt{escalate}\) otherwise. Therefore, the corresponding CRC loss on an example \((x,y)\) is defined as
\begin{equation}
L_d(\tau; x,y)
=
\mathbbm{1}\!\Big[r_{\theta,d}\!\big(\tilde u_d(x)\big)\le \tau\Big]\,
\mathbbm{1}\!\Big[\hat y_d(x)\neq y_d\Big].
\label{eq:crc_loss}
\end{equation}
This is exactly the indicator of a wrong-accept event. Accordingly, for deployment-time inputs and labels \((X,Y)\), the threshold-induced per-finding wrong-accept risk can be written as \(R_d(\tau)=\mathbb{E}[L_d(\tau;X,Y)]\), which is the threshold-parameterized version of the risk \(R_d\) in Sec.~\ref{subsec:problem_formulation}. The same threshold also determines the corresponding coverage, thereby linking threshold calibration directly to the risk--coverage objective.

The threshold-dependent wrong-accept loss \(L_d(\tau; x,y)\) provides the quantity on which CRC calibrates \(\tau\) into a deployable threshold \(\hat\tau_d\) that satisfies the target risk level. In deployment settings, the user specifies a target risk level \(\alpha\in(0,1)\) according to the desired tolerance for risks. CRC then calibrates a per-finding acceptance threshold \(\hat\tau_d\) on a separate calibration split such that, under the standard exchangeability assumption between calibration and deployment samples, the induced expected wrong-accept risk is controlled:
\begin{equation}
\mathbb{E}\!\Big[L_d(\hat\tau_d;X,Y)\Big]\le \alpha.
\label{eq:crc_guarantee}
\end{equation}
The calibrated threshold \(\hat\tau_d\) is then used at deployment to instantiate the per-finding accept-versus-escalate routing rule:
\begin{equation}
a_d(x)=
\begin{cases}
\texttt{accept}, & r_{\theta,d}\!\big(\tilde u_d(x)\big)\le \hat\tau_d,\\
\texttt{escalate}, & r_{\theta,d}\!\big(\tilde u_d(x)\big)> \hat\tau_d,
\end{cases}
\label{eq:final_policy}
\end{equation}
where \(a_d(x)\) denotes the router action for finding \(d\). 

Taken together, our proposed method first constructs a per-finding multi-signal uncertainty representation, then maps it to a learned risk score, and finally uses CRC to calibrate a deployment threshold that enforces the target wrong-accept risk while preserving as much autonomous coverage as possible. \textbf{This yields a risk-constrained, uncertainty-aware, and model-agnostic routing module that can be attached to a broad range of predictive or agentic medical systems without modifying the underlying predictor(s).}

\begin{table*}[t]
\centering
\caption{Per-disease risk ($\hat{R}_d$) across routing methods ($\alpha = 5\%$). All selective methods are matched to ${\sim}$91\% macro coverage. Best results among selective methods are highlighted in bold. \checkmark\ indicates $\hat{R}_d \leq \alpha$. \textsuperscript{$\dagger$}Relative reduction of CRC-Router risk compared to Accept-All; ``--'' indicates the Accept-All risk is already below $\alpha$.}
\label{tab:main_results}
\begin{tabular*}{\textwidth}{@{\extracolsep{\fill}}lcccccc@{}}
\toprule
Disease & \makecell{Accept\\-All} & \makecell{MaxProb\\~\cite{jaeger2022call}} & \makecell{MCD-MI\\~\cite{traub2024overcoming}} & \makecell{Conformal\\Pred.~\cite{angelopoulos2023conformal}} & \textbf{CRC-Router} & \makecell{Risk\\Red.\textsuperscript{$\dagger$}} \\
\midrule
Atelectasis         & 17.0\% & 5.4\%          & 10.6\%          & 11.9\%          & 3.9\%\checkmark & $\downarrow$77\% \\
Cardiomegaly        & 3.8\%\checkmark  & 1.5\%\checkmark & 1.3\%\checkmark & 1.1\%\checkmark & 3.8\%\checkmark & --      \\
Consolidation       & 18.1\% & 9.8\%          & 17.8\%          & 9.6\%          & 4.4\%\checkmark & $\downarrow$76\% \\
Edema               & 2.8\%\checkmark  & 0.8\%\checkmark & 0.8\%\checkmark & 1.0\%\checkmark & 2.8\%\checkmark & --      \\
Effusion            & 14.8\% & 5.1\%          & 9.3\%          & 9.6\%          & 4.4\%\checkmark & $\downarrow$70\% \\
Emphysema           & 2.3\%\checkmark  & 1.4\%\checkmark & 0.8\%\checkmark & 0.9\%\checkmark & 2.3\%\checkmark & --      \\
Fibrosis            & 6.4\%  & 4.8\%\checkmark & 3.8\%\checkmark & 1.2\%\checkmark & 5.0\%\checkmark & $\downarrow$23\% \\
Hernia              & 1.1\%\checkmark  & 1.1\%\checkmark & 1.1\%\checkmark & 0.1\%\checkmark & 1.1\%\checkmark & --      \\
Infiltration        & 21.9\% & 7.4\%          & 19.7\%          & 16.7\%          & 3.5\%\checkmark & $\downarrow$84\% \\
Mass                & 7.3\%  & 4.2\%\checkmark & 3.6\%\checkmark & 3.3\%\checkmark & 4.5\%\checkmark & $\downarrow$38\% \\
Nodule              & 8.5\%  & 6.3\%          & 5.9\%          & 3.9\%\checkmark & 4.5\%\checkmark & $\downarrow$47\% \\
Pleural Thickening  & 8.2\%  & 5.1\%          & 5.2\%          & 2.0\%\checkmark & 4.3\%\checkmark & $\downarrow$47\% \\
Pneumonia           & 9.3\%  & 7.5\%          & 8.2\% & 1.3\%\checkmark & 4.6\%\checkmark & $\downarrow$50\% \\
Pneumothorax        & 6.8\%  & 2.7\%\checkmark & 2.4\%\checkmark          & 2.7\%\checkmark & 5.9\%          & $\downarrow$13\% \\
\midrule
Macro               & 9.2\%  & 4.5\%          & 6.5\%          & 4.7\%          & \textbf{3.9\%} & $\downarrow$58\% \\
Coverage            & 100.0\% & 91.0\%         & 91.1\%         & 91.1\%         & \textbf{91.3\%} & --      \\
Pass ($\leq$5\%)    & 4/14   & 7/14           & 7/14           & 10/14          & \textbf{13/14} & --      \\
\bottomrule
\end{tabular*}

\end{table*}

\section{Experiments}

\begin{table*}[t]
\centering
\caption{Per-disease wrong-accept risk ($\hat{R}_d$) for MedRAX integration at $\alpha=5\%$ on NIH ChestX-ray14 and CheXpert. CR\textsuperscript{*}=CRC-Router. MoE is omitted on CheXpert as the committee was not trained on CheXpert. \checkmark\ marks $\hat{R}_d \le \alpha$.}
\label{tab:ablation_conditions}
\begin{tabular*}{\textwidth}{@{\extracolsep{\fill}}lccc|lcc@{}}
\hline
\multicolumn{4}{c|}{NIH ChestX-ray14~\cite{wang2017chestx}} & \multicolumn{3}{c}{CheXpert~\cite{irvin2019chexpert}} \\
\hline
Disease & MedRAX~\cite{fallahpour2025medrax} & \makecell{+CR\textsuperscript{*}\\(TTA)} & \makecell{+CR\textsuperscript{*}\\(MoE)} & Disease & MedRAX~\cite{fallahpour2025medrax} & \makecell{+CR\textsuperscript{*}\\(TTA)} \\
\hline
Atelectasis         & 6.6\%          & 4.7\%\checkmark  & 4.0\%\checkmark & Enl.\ Cardiomed.    & 57.6\%         & 4.2\%\checkmark \\
Cardiomegaly        & 12.9\%         & 3.6\%\checkmark  & 3.7\%\checkmark & Cardiomegaly        & 51.3\%         & 4.7\%\checkmark \\
Consolidation       & 9.6\%          & 6.6\%            & 3.4\%\checkmark & Lung Opacity        & 30.1\%         & 4.7\%\checkmark \\
Edema               & 6.5\%          & 6.2\%            & 2.9\%\checkmark & Lung Lesion         & 1.5\%\checkmark & 3.9\%\checkmark \\
Effusion            & 14.5\%         & 5.9\%            & 4.3\%\checkmark & Edema               & 19.4\%         & 4.3\%\checkmark \\
Emphysema           & 1.6\%\checkmark & 2.6\%\checkmark & 1.7\%\checkmark & Consolidation       & 19.3\%         & 5.4\% \\
Fibrosis            & 2.7\%\checkmark & 4.3\%\checkmark & 4.5\%\checkmark & Pneumonia           & 1.4\%\checkmark & 6.3\% \\
Hernia              & 0.5\%\checkmark & 0.4\%\checkmark & 1.5\%\checkmark & Atelectasis         & 33.6\%         & 6.2\% \\
Infiltration        & 9.4\%          & 5.6\%            & 3.7\%\checkmark & Pneumothorax        & 11.6\%         & 4.4\%\checkmark \\
Mass                & 1.6\%\checkmark & 5.6\%           & 3.7\%\checkmark & Pleural Effusion    & 22.3\%         & 4.2\%\checkmark \\
Nodule              & 2.0\%\checkmark & 7.0\%           & 4.7\%\checkmark & Pleural Other       & 0.2\%\checkmark & 1.5\%\checkmark \\
Pl.\ Thickening     & 1.5\%\checkmark & 4.2\%\checkmark & 3.7\%\checkmark & Fracture            & 3.2\%\checkmark & 4.8\%\checkmark \\
Pneumonia           & 4.1\%\checkmark & 3.2\%\checkmark & 4.8\%\checkmark &                     &                &  \\
Pneumothorax        & 6.2\%          & 6.9\%            & 6.8\%           &                     &                &  \\
\hline
Macro Risk          & 5.7\%          & 4.8\%            & \textbf{3.8\%} & Macro Risk & 20.9\% & \textbf{4.6\%} \\
Coverage            & 62.9\%         & 81.5\%           & \textbf{91.3\%}          & Coverage   & 65.6\% & \textbf{68.4\%} \\
Pass ($\leq$5\%)    & 7/14           & 7/14             & \textbf{13/14}           & Pass ($\leq$5\%) & 4/12 & \textbf{9/12} \\
\hline
\end{tabular*}

\end{table*}

\begin{figure*}[t]
  \centering
  \includegraphics[width=\textwidth]{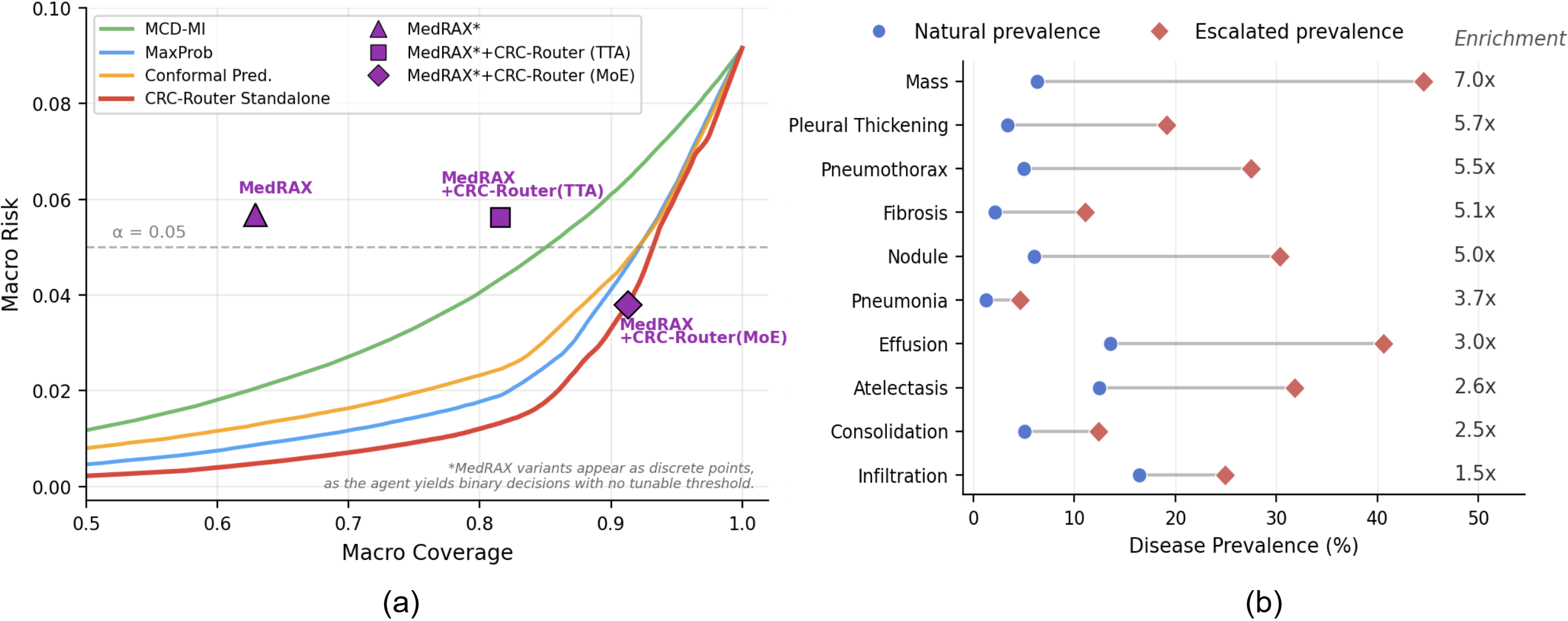}
  \caption{(a) Risk--Coverage Tradeoff curves for CRC-Router and baselines on NIH ChestX-ray14. (b) Prevalence enrichment among escalated cases. }
  \label{fig:enrichment}
\end{figure*}

We first evaluate CRC-Router as a standalone risk-constrained router on NIH ChestX-ray14~\cite{wang2017chestx} (112,120 frontal-view CXRs; 14 labels). Following standard practice, we use a \textbf{patient-level} Train/Calibration/Test split of 0.5/0.25/0.25, and split Calibration equally into risk-model training and CRC calibration subsets. To estimate epistemic and aleatoric uncertainty, we train an MoE committee of five multi-label classifiers (RegNetY-1.6~\cite{radosavovic2020designing}, ResNet-50d~\cite{he2016deep}, RexNet-150~\cite{han2021rethinking}, EfficientNetV2-B3~\cite{tan2021efficientnetv2}, and DenseNet-121~\cite{huang2017densely}) with learning rate \(10^{-4}\), batch size 24, and early stopping (patience 10; max 60 epochs). The Mahalanobis score is computed from 24 handcrafted radiomics features (13 intensity, 5 GLCM texture, 6 histogram) using a diagonal Mahalanobis model fitted on the training set. Per-disease risk models \(r_{\theta,d}\) are selected by grid search over MLP architectures using 5-fold cross-validated AUROC. CRC thresholds \(\hat\tau_d\) are calibrated per disease at target risk \(\alpha=5\%\). We compare against standard post-hoc selective prediction baselines: maximum predictive probability thresholding (MaxProb)~\cite{jaeger2022call}, MCD-MI (computed via ensemble disagreement)~\cite{traub2024overcoming}, and Conformal Prediction~\cite{angelopoulos2023conformal}. Baseline thresholds are selected to approximately match CRC-Router macro coverage, and risk is evaluated at the matched operating point.

To evaluate adaptability to state-of-the-art medical AI agents, we further integrate CRC-Router into MedRAX~\cite{fallahpour2025medrax}, a representative chest X-ray agentic AI system, and report results on both NIH ChestX-ray14 and CheXpert~\cite{irvin2019chexpert}. For computational efficiency, we use stratified subsampling: on NIH ChestX-ray14 we evaluate on a 10\% stratified test subsample, and on CheXpert we use stratified subsamples of 2,000 images each for risk-model training, CRC calibration, and testing (U-Ignore policy~\cite{irvin2019chexpert}, 12 pathology labels, excluding No Finding and Support Devices). We evaluate three conditions when applicable: (1) MedRAX, where the LLM decides accept versus escalate without CRC-Router guidance; (2) MedRAX + CRC-Router (TTA), where CRC-Router is calibrated on MedRAX's built-in DenseNet and epistemic/aleatoric uncertainty is estimated via test-time augmentation (TTA) by sampling radiology-safe perturbations (rotation \(\pm5\degree\), translation \(\pm2\%\), scaling \(\pm2\%\), brightness jitter \(\pm5\%\)) to generate 5 stochastic forward passes per image, testing plug-in transfer when no ensemble committee is available; (3) MedRAX + CRC-Router (MoE) on NIH ChestX-ray14 only, where MedRAX's DenseNet is replaced by our MoE committee with CRC-Router.

We primarily evaluate all methods with risk-coverage analysis. For each disease \(d\), we report empirical wrong-accept risk \(\widehat{R}_d = n^{\mathrm{wrong}}_d / n^{\mathrm{total}}_d\), where \(n^{\mathrm{wrong}}_d\) is the number of accepted but incorrect predictions and \(n^{\mathrm{total}}_d\) is the total number of test cases for disease \(d\). We additionally report per-disease coverage and the number of diseases with \(\widehat{R}_d \le \alpha\). We set \(\alpha=5\%\) in all experiments. 

\section{Results and Discussion}

\subsubsection{CRC-Router Standalone Results}
As shown in Table~\ref{tab:main_results}, CRC-Router achieves the lowest macro wrong-accept risk among all selective routing methods at matched macro coverage, with a macro risk of \textbf{3.9\%} at \textbf{91.3\%} coverage. Although Accept-All attains 100\% coverage by definition, it yields a substantially higher macro wrong-accept risk of 9.2\%. Compared with Accept-All, CRC-Router reduces macro wrong-accept risk by \textbf{58\%} while sacrificing only \textbf{8.7 percentage points} of coverage, demonstrating a favorable safety--automation trade-off. It also shows the strongest empirical conformity to the target risk level (\(\alpha=5\%\)): \textbf{13/14} diseases achieve per-disease wrong-accept risk \(\leq 5\%\), compared with 10/14 for Conformal Prediction, and 7/14 for both MaxProb and Entropy baselines. The only disease above the target is Pneumothorax (5.9\%), which represents a modest deviation from the nominal level on the held-out test set. 
Fig.~\ref{fig:enrichment}(a) further illustrates the risk--coverage trade-off across routing methods. CRC-Router provides the most favorable trade-off in the high-coverage regime, consistently achieving lower macro risk than baselines at comparable coverage levels.

For the downstream CXR triage task, Fig.~\ref{fig:enrichment}(b) shows that the proposed router substantially enriches pathology prevalence in the escalated queue. Across the 10 diseases with non-trivial escalation rates, the prevalence among escalated cases increases to 1.5--7.0$\times$ the natural population prevalence. These results indicate that escalated cases are concentrated in diagnostically challenging and clinically important subsets, which supports the use of CRC-Router as a safety-oriented triage mechanism for downstream review in practical clinical workflows. The four diseases with near-100\% acceptance (Cardiomegaly, Edema, Emphysema, and Hernia) are omitted from Fig.~\ref{fig:enrichment}(b) because they produce negligible escalation queues.

\begin{table}[t]
\centering
\small
\setlength{\tabcolsep}{3pt}
\caption{Ablation study on features in uncertainty vector. $\Delta$ Cov.\ denotes the change in macro coverage relative to the full model.}
\label{tab:ablation_feature_importance}
\begin{tabular*}{\columnwidth}{@{\extracolsep{\fill}}lcc@{}}
\toprule
Removed Feature & Macro Cov. & $\Delta$ Cov. \\
\midrule
None (full model)   & \textbf{91.3}\% & ---         \\
${H}_{\bar{p}}$     & 90.4\% & $-$0.9\%    \\
Mahalanobis         & 90.0\% & $-$1.3\%    \\
Epistemic           & 88.2\% & $-$3.1\%    \\
Aleatoric           & 90.3\% & $-$1.0\%    \\
$\bar{p}$           & 85.1\% & $-$6.2\%    \\
\bottomrule
\end{tabular*}
\end{table}

\subsubsection{MedRAX Integration Results}
Table~\ref{tab:ablation_conditions} reports per-disease wrong-accept risk (\(\hat R_d\)) for MedRAX integration on NIH ChestX-ray14 and CheXpert. On NIH ChestX-ray14, MedRAX alone achieves 62.9\% macro coverage at 5.7\% macro risk, while integrating CRC-Router (TTA) substantially improves the risk--coverage trade-off to 81.5\% coverage and 4.8\% risk. On CheXpert, CRC-Router (TTA) similarly reduces macro risk from 20.9\% to 4.6\% while slightly increasing coverage from 65.6\% to 68.4\%. This demonstrates that CRC-Router can serve as a modular plug-in routing component for an existing agentic framework without retraining or modifying the agent architecture, and it remains effective when epistemic and aleatoric signals are approximated via TTA rather than an ensemble. On NIH ChestX-ray14, using CRC-Router (MoE) yields the best performance, reaching 91.3\% coverage at 3.8\% risk. Overall, these results indicate complementary benefits from CRC-based risk-constrained routing and stronger uncertainty estimates, and support the cross-dataset generalizability of CRC-Router when embedded within a medical agentic workflow.

\subsubsection{Ablation Study}
To quantify the contribution of each uncertainty signal in the proposed uncertainty vector, we conduct a leave-one-feature-out ablation by removing one component at a time and retraining all 14 per-disease risk models using the same protocol and architecture. Table~\ref{tab:ablation_feature_importance} shows that the full model attains the highest macro coverage, 91.3\%, at the target risk level. Removing any individual component decreases macro coverage, indicating that each feature contributes complementary information for risk estimation. Among the uncertainty terms, removing the epistemic component causes the largest degradation, with macro coverage \(\downarrow 3.1\%\), suggesting that model-disagreement information is particularly important for identifying high-risk cases. Across all features, removing the predictive score term \(\bar p\) produces the largest overall reduction, with macro coverage \(\downarrow 6.2\%\), indicating that prediction magnitude remains an important complement to uncertainty features when learning the uncertainty-to-risk mapping.

\section{Conclusions}


In this paper, we presented CRC-Router, a risk-constrained, uncertainty-aware routing module for both conventional medical prediction models and agentic medical AI systems. CRC-Router combines heterogeneous uncertainty signals and the predictive score to estimate per-finding wrong-accept risk, and then applies Conformal Risk Control (CRC) to calibrate deployment thresholds for a user-specified risk target. Using CXR multi-finding triage as a case study, CRC-Router was effective both as a standalone routing layer and as a plug-in module integrated with MedRAX. Across both settings, it achieved the best empirical risk--coverage trade-off among compared methods, supporting its use for risk-constrained medical AI deployment and highlighting its modular, model-agnostic compatibility with existing predictive and agentic medical AI systems.



\bibliographystyle{IEEEtran}
\bibliography{bibliography}

\begin{thebibliography}{10}
\providecommand{\url}[1]{#1}
\csname url@samestyle\endcsname
\providecommand{\newblock}{\relax}
\providecommand{\bibinfo}[2]{#2}
\providecommand{\BIBentrySTDinterwordspacing}{\spaceskip=0pt\relax}
\providecommand{\BIBentryALTinterwordstretchfactor}{4}
\providecommand{\BIBentryALTinterwordspacing}{\spaceskip=\fontdimen2\font plus
\BIBentryALTinterwordstretchfactor\fontdimen3\font minus \fontdimen4\font\relax}
\providecommand{\BIBforeignlanguage}[2]{{%
\expandafter\ifx\csname l@#1\endcsname\relax
\typeout{** WARNING: IEEEtran.bst: No hyphenation pattern has been}%
\typeout{** loaded for the language `#1'. Using the pattern for}%
\typeout{** the default language instead.}%
\else
\language=\csname l@#1\endcsname
\fi
#2}}
\providecommand{\BIBdecl}{\relax}
\BIBdecl

\bibitem{xu2025comprehensive}
G.~Xu, X.~Li, Y.~Chen, Y.~Duan, S.~Wu, A.~Yu, C.-H. Chiu, J.~Ni, N.~Tang, T.~J.-J. Li \emph{et~al.}, ``A comprehensive survey of agentic {AI} in healthcare,'' \emph{Authorea Preprints}, 2025.

\bibitem{han2024towards}
J.~Han, W.~Buntine, and E.~Shareghi, ``Towards uncertainty-aware language agent,'' in \emph{Findings of the Association for Computational Linguistics: ACL 2024}, 2024, pp. 6662--6685.

\bibitem{zhao2025uncertainty}
Q.~Zhao, D.~Li, Y.~Liu, W.~Cheng, Y.~Sun, M.~Oishi, T.~Osaki, K.~Matsuda, H.~Yao, C.~Zhao \emph{et~al.}, ``Uncertainty propagation on {LLM} agent,'' in \emph{Proceedings of the 63rd Annual Meeting of the Association for Computational Linguistics (Volume 1: Long Papers)}, 2025, pp. 6064--6073.

\bibitem{zhi2025seeing}
Z.~Zhi, C.~Feng, A.~Daneshmend, M.~Orlu, A.~Demosthenous, L.~Yin, D.~Li, Z.~Liu, and M.~R. Rodrigues, ``Seeing and reasoning with confidence: Supercharging multimodal {LLMs} with an uncertainty-aware agentic framework,'' \emph{arXiv preprint arXiv:2503.08308}, 2025.

\bibitem{zhang2026agentic}
J.~Zhang, P.~K. Choubey, K.-H. Huang, C.~Xiong, and C.-S. Wu, ``Agentic uncertainty quantification,'' \emph{arXiv preprint arXiv:2601.15703}, 2026.

\bibitem{angelopoulos2022conformal}
A.~N. Angelopoulos, S.~Bates, A.~Fisch, L.~Lei, and T.~Schuster, ``Conformal risk control,'' \emph{arXiv preprint arXiv:2208.02814}, 2022.

\bibitem{silva2025trustworthy}
J.~Silva-Rodr{\'\i}guez, I.~Ben~Ayed, and J.~Dolz, ``Trustworthy few-shot transfer of medical {VLMs} through split conformal prediction,'' in \emph{International Conference on Medical Image Computing and Computer-Assisted Intervention}.\hskip 1em plus 0.5em minus 0.4em\relax Springer, 2025, pp. 658--668.

\bibitem{teneggi2025conformal}
J.~Teneggi, J.~W. Stayman, and J.~Sulam, ``Conformal risk control for semantic uncertainty quantification in computed tomography,'' in \emph{International Conference on Medical Image Computing and Computer-Assisted Intervention}.\hskip 1em plus 0.5em minus 0.4em\relax Springer, 2025, pp. 45--55.

\bibitem{geifman2017selective}
Y.~Geifman and R.~El-Yaniv, ``Selective classification for deep neural networks,'' \emph{Advances in neural information processing systems}, vol.~30, 2017.

\bibitem{fallahpour2025medrax}
A.~Fallahpour, J.~Ma, A.~Munim, H.~Lyu, and B.~Wang, ``{MedRAX}: Medical reasoning agent for chest {X-ray},'' \emph{arXiv preprint arXiv:2502.02673}, 2025.

\bibitem{Kim2024MDAgents}
Y.~Kim, C.~Park, H.~Jeong, Y.~S. Chan, X.~Xu, D.~McDuff, H.~Lee, M.~Ghassemi, C.~Breazeal, and H.~W. Park, ``{MDAgents}: An adaptive collaboration of {LLMs} for medical decision-making,'' \emph{Advances in Neural Information Processing Systems}, vol.~37, pp. 79\,410--79\,452, 2024.

\bibitem{Li2024MMedAgent}
B.~Li, T.~Yan, Y.~Pan, J.~Luo, R.~Ji, J.~Ding, Z.~Xu, S.~Liu, H.~Dong, Z.~Lin \emph{et~al.}, ``{MMedAgent}: Learning to use medical tools with multi-modal agent,'' \emph{arXiv preprint arXiv:2407.02483}, 2024.

\bibitem{Chen2025RadFabric}
W.~Chen, Y.~Dong, Z.~Ding, Y.~Shi, Y.~Zhou, F.~Zeng, Y.~Luo, T.~Lin, Y.~Su, Y.~Wu, K.~Zhang, Z.~Xiang, T.~Liu, N.~Liu, L.~Sun, Y.~Yuan, and X.~Li, ``Radfabric: Agentic ai system with reasoning capability for radiology,'' \emph{arXiv preprint arXiv:2506.14142}, 2025.

\bibitem{Guo2017Calibration}
C.~Guo, G.~Pleiss, Y.~Sun, and K.~Q. Weinberger, ``On calibration of modern neural networks,'' in \emph{Proceedings of the 34th International Conference on Machine Learning}, ser. Proceedings of Machine Learning Research, vol.~70, 2017, pp. 1321--1330.

\bibitem{GalGhahramani2016}
Y.~Gal and Z.~Ghahramani, ``Dropout as a bayesian approximation: Representing model uncertainty in deep learning,'' \emph{arXiv preprint arXiv:1506.02142}, 2016.

\bibitem{KendallGal2017}
A.~Kendall and Y.~Gal, ``What uncertainties do we need in bayesian deep learning for computer vision?'' in \emph{Advances in Neural Information Processing Systems}, vol.~30, 2017.

\bibitem{Ovadia2019}
Y.~Ovadia, E.~Fertig, J.~Ren, Z.~Nado, D.~Sculley, S.~Nowozin, J.~V. Dillon, B.~Lakshminarayanan, and J.~Snoek, ``Can you trust your model's uncertainty? evaluating predictive uncertainty under dataset shift,'' in \emph{Advances in Neural Information Processing Systems}, vol.~32, 2019.

\bibitem{Lee2018Mahalanobis}
K.~Lee, K.~Lee, H.~Lee, and J.~Shin, ``A simple unified framework for detecting out-of-distribution samples and adversarial attacks,'' in \emph{Advances in Neural Information Processing Systems}, vol.~31, 2018.

\bibitem{angelopoulos2023conformal}
A.~N. Angelopoulos and S.~Bates, ``Conformal prediction: A gentle introduction,'' \emph{Foundations and Trends in Machine Learning}, vol.~16, no.~4, pp. 494--591, 2023.

\bibitem{jaeger2022call}
P.~F. Jaeger, C.~T. L{\"u}th, L.~Klein, and T.~J. Bungert, ``A call to reflect on evaluation practices for failure detection in image classification,'' \emph{arXiv preprint arXiv:2211.15259}, 2022.

\bibitem{traub2024overcoming}
J.~Traub, T.~J. Bungert, C.~T. L{\"u}th, M.~Baumgartner, K.~H. Maier-Hein, L.~Maier-Hein, and P.~F. J{\"a}ger, ``Overcoming common flaws in the evaluation of selective classification systems,'' \emph{Advances in Neural Information Processing Systems}, vol.~37, pp. 2323--2347, 2024.

\bibitem{wang2017chestx}
X.~Wang, Y.~Peng, L.~Lu, Z.~Lu, M.~Bagheri, and R.~M. Summers, ``{ChestX-Ray8}: Hospital-scale chest {X}-ray database and benchmarks on weakly-supervised classification and localization of common thorax diseases,'' in \emph{Proceedings of the IEEE Conference on Computer Vision and Pattern Recognition}, 2017, pp. 2097--2106.

\bibitem{irvin2019chexpert}
J.~Irvin, P.~Rajpurkar, M.~Ko, Y.~Yu, S.~Ciurea-Ilcus, C.~Chute, H.~Marklund, B.~Haghgoo, R.~Ball, K.~Shpanskaya \emph{et~al.}, ``Chexpert: A large chest radiograph dataset with uncertainty labels and expert comparison,'' in \emph{Proceedings of the AAAI conference on artificial intelligence}, vol.~33, no.~01, 2019, pp. 590--597.

\bibitem{radosavovic2020designing}
I.~Radosavovic, R.~P. Kosaraju, R.~Girshick, K.~He, and P.~Doll{\'a}r, ``Designing network design spaces,'' in \emph{Proceedings of the IEEE/CVF Conference on Computer Vision and Pattern Recognition}, 2020, pp. 10\,428--10\,436.

\bibitem{he2016deep}
K.~He, X.~Zhang, S.~Ren, and J.~Sun, ``Deep residual learning for image recognition,'' in \emph{Proceedings of the IEEE Conference on Computer Vision and Pattern Recognition}, 2016, pp. 770--778.

\bibitem{han2021rethinking}
D.~Han, S.~Yun, B.~Heo, and Y.~Yoo, ``Rethinking channel dimensions for efficient model design,'' in \emph{Proceedings of the IEEE/CVF Conference on Computer Vision and Pattern Recognition}, 2021, pp. 732--741.

\bibitem{tan2021efficientnetv2}
M.~Tan and Q.~Le, ``{EfficientNetV2}: Smaller models and faster training,'' in \emph{International Conference on Machine Learning}.\hskip 1em plus 0.5em minus 0.4em\relax PMLR, 2021, pp. 10\,096--10\,106.

\bibitem{huang2017densely}
G.~Huang, Z.~Liu, L.~Van Der~Maaten, and K.~Q. Weinberger, ``Densely connected convolutional networks,'' in \emph{Proceedings of the IEEE Conference on Computer Vision and Pattern Recognition}, 2017, pp. 4700--4708.

\end{thebibliography}

\end{document}